\documentclass[10pt]{article}
\usepackage[a4paper, total={5in, 8in}]{geometry}
\usepackage{times}
\usepackage{epsfig}

\usepackage[utf8]{inputenc}
\usepackage{graphicx,verbatimbox}
\usepackage{tcolorbox}

\usepackage{rotating}
\usepackage{palatino}

\usepackage{tikz}
\usepackage{comment}
\usepackage{amsmath,amssymb} %

\usepackage{booktabs} %
\usepackage{xcolor}
\usepackage{color,colortbl}
\usepackage{pifont}
\usepackage[colorlinks=true,linkcolor=blue, citecolor=cyan]{hyperref}%

\usepackage{multirow}
\usepackage[font=footnotesize]{caption}
\usepackage{placeins}

\usepackage[accsupp]{axessibility}  %

\usepackage{booktabs,tabularx}
\usepackage{multirow}
\usepackage[numbers,sort&compress]{natbib}
\usepackage[colorlinks=true,linkcolor=blue, citecolor=cyan]{hyperref}%
\usepackage{pifont}
\usepackage{xspace}
\usepackage{algorithm}
\usepackage{listings}

\newcommand{\cmark}{\textcolor{blue}{\ding{51}\xspace}}%
\newcommand{\xmark}{\textcolor{red}{\ding{55}\xspace}}%
\newcommand{\xmarkg}{\textcolor{lightgray}{\ding{55}}\xspace}%

\definecolor{Goldenrod}{RGB}{245,245,220}

\def\mypar#1{\medskip{\noindent\bf #1}\hspace{1mm}}

\def \pzo {\phantom{0}}

\newcommand{\stdminus}[1]{\scalebox{0.65}{$\pm #1$}}

\title{Three things everyone should know\\
about Vision Transformers} 
\author{
\begin{minipage}{\linewidth}
\begin{center}
\normalsize Hugo Touvron$^{\star,\dagger}$ \hspace{0.35cm} Matthieu Cord$^{\dagger}$ \hspace{0.35cm} Alaaeldin El-Nouby$^{\star,\diamond}$  \\[0.2cm]
Jakob Verbeek$^{\star}$ \quad \quad Herv\'e J\'egou$^{\star}$ \\[0.5cm] 
\scalebox{1.}{$^\star$Meta AI\hspace{0.6cm} $^\dagger$Sorbonne University \hspace{0.6cm} $^\diamond$Inria}\\[1cm]
\end{center}
\end{minipage}
}
\date{~}

\begin{document}

\maketitle

\begin{abstract}
After their initial success in natural language processing, transformer architectures have rapidly gained traction in computer vision, providing state-of-the-art results for tasks such as image classification, detection, segmentation, and video analysis.
We offer three insights based on simple and easy to implement variants of vision transformers. 
(1) The residual layers of vision transformers, which are usually processed sequentially, can to some extent be processed efficiently in parallel without noticeably affecting the accuracy. 
(2) Fine-tuning the weights of the attention layers is sufficient to adapt vision transformers to a higher resolution and to other classification tasks. 
This saves compute, reduces the peak memory consumption at fine-tuning time, and allows sharing the majority of weights across tasks.
(3) Adding  MLP-based   patch pre-processing layers improves Bert-like self-supervised training based on patch masking. 
We evaluate the impact of these design choices using the ImageNet-1k dataset, and confirm our findings on the ImageNet-v2 test set. 
Transfer performance is measured across six smaller datasets.

\end{abstract}

\section{Introduction}

Since its introduction the Transformer architecture~\cite{vaswani2017attention}  has become the dominant architecture in natural language processing tasks, replacing previously popular recurrent architectures. 
The vision transformer~\cite{dosovitskiy2020image} (ViT) is a simple adaptation of transformers to  computer vision tasks like image classification: the input  image is divided  into non-overlapping patches, which are fed to a vanilla transformer architecture, after a linear patch projection layer.
In contrast to  networks built from convolutional layers, transformers offer parallel processing and a complete field-of-view in a single layer.
Along with other attention-based architectures, see e.g.~\cite{bello2019attention,carion2020end}, transformers have recently substantially influenced the design of computer vision architectures. 
Many modern architectures in computer vision directly inherit parts of their design from this work, or are at least inspired by the recent findings resulting from transformers~\cite{carion2020end,dosovitskiy2020image,Touvron2020TrainingDI}. 
As a result, significant improvements have been observed on different computer vision tasks, ranging from object detection and segmentation~\cite{el2021xcit} and video analysis~\cite{fan2021multiscale,arnab2021vivit} to image generation~\cite{chang2022maskgit,hudson21icml}. 

While vision transformers have led to considerable progress, the optimization of their design and training procedures have only been explored to a limited extent.
In this paper, we offer three insights on training vision transformers. 

\mypar{1. Parallel vision transformers.} Several works~\cite{Zagoruyko2016WideRN,Goyal2021NondeepN}  advocate the interest shallower networks for reasons ranging from lower latency to easier optimization. 
We propose a very simple way to achieve this with ViTs. Let us denote by MHSA the multi-headed self-attention residual block, and by FFN the residual feedforward network. Starting from a sequential architecture depicted as follows,

\begin{figure}
    \centering
    \includegraphics[trim={0 1.5cm 0 1.5cm}, scale=0.4]{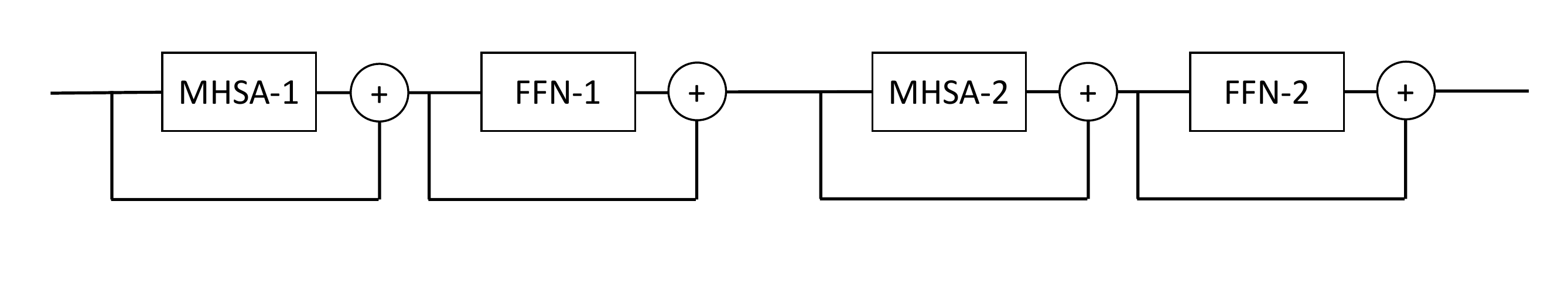}
\end{figure}
\noindent we parallelize the architecture by reorganizing the same blocks by pairs,

\begin{figure}
    \centering
    \includegraphics[trim={0 1.5cm 0 1.5cm},scale=0.4]{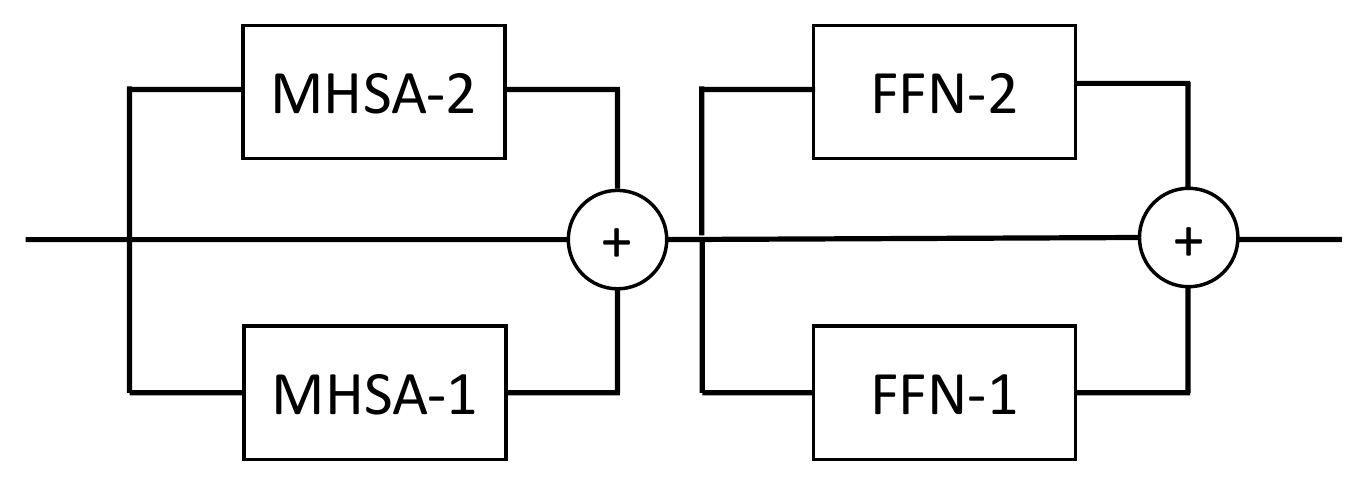}
    \label{fig:my_label}
\end{figure}

\noindent which can be done for any different numbers of parallel blocks. 
This produces an architecture with the same number of parameters and compute, while being  wider and shallower. 
This design allows for more parallel processing, easing optimization and reducing latency depending on the implementation.

In Section~\ref{sec:seq_vs_par}, we experimentally analyse the performance of this parallel construction, and in particular how it affects the accuracy in comparison to the sequential baseline. The parallel version  becomes a compelling option if deep enough. In some cases, we observe improvements in accuracy resulting from an easier optimization. Regarding the latency on  GPUs, we observe %
reductions %
in the case of %
small batch sizes.\footnote{We have not found any papers in the literature analyzing the effect of width versus depth for ViT on common GPUs and CPUs. %
}

\mypar{2. Fine-tuning attention is all you need.} 
It is common practice to pre-train networks before fine-tuning them on a target task. 
This is the standard approach underpinning transfer learning, where one leverages a large generic dataset like ImageNet~\cite{Russakovsky2015ImageNet12} when the number of images is limited for the target task \cite{oquab2014learning,Yosinski2014HowTA}. 
Another context is the one of changing resolution. Typically one would train at a lower resolution than the one employed at inference time. This saves resources, but additionally it reduces the discrepancy of scale between train and test images that results from data augmentation~\cite{Touvron2019FixRes}. 
In Section~\ref{sec:finetune} we show that, in the case of ViT, it is mostly sufficient to fine-tune only the multi-head attention layers and freeze the feedforward network (FFN) layers. This saves compute and reduces the memory peak during training. 
Importantly this allows the same FFN weights, which dominate the number of parameters, to be used for multiple tasks. The impact on accuracy is statistically not significant when fine-tuning for different image resolutions. For large models, the impact on accuracy  is limited when considering transfer to other classification tasks. 

\mypar{3. Patch preprocessing with masked self-supervised learning.} 
The first layers of a transformer have a relatively local span~\cite{dAscoli2021ConViTIV}, suggesting that they mostly behave like convolutions. 
Some recent hybrid architectures~\cite{graham2021levit,el2021xcit,Han2021TransformerIT} preprocess their input images with a convolutional stem, to improve accuracy and training stability~\cite{Xiao2021EarlyCH}. 
However, preprocessing images with convolutions is \textit{a priori} not compatible with the recent and successful mask-based self-supervised learning approaches, like BeiT~\cite{bao2021beit} or  MAE~\cite{he2021masked}. 
The convolutions propagate information across patches, impeding the masked prediction task. 

In  Section~\ref{sec:stem}, we propose a simple way to adapt mask-based self-supervised training methods with patch pre-processing, by applying the masking after the patch pre-processing. 
However, our analysis reveals that existing convolutional stems are not effective when combined with BeiT. To address this issue, we introduce a hierarchical MLP (hMLP) stem that interleaves MLP layers and patch aggregation operations, and prohibits any communication between patches. Our experiments show that this choice is effective and able to leverage the benefit of both BeiT self-supervised pre-training and patch pre-processing. 
Moreover, our hMLP-stem  is also effective for ViT in the supervised case: it is on par with the best convolutional stem of our comparison~\cite{graham2021levit}. 

\section{Background} 

In this section, we discuss  related work in common with  our different contributions. 
We also introduce the baseline  ViT models considered in this study and  how they are trained. 
In subsequent sections, we discuss related work that is  more specific  to each of our three specific contributions.  

\subsection{Related work} 
Attention-based models, and in particular transformers~\cite{vaswani2017attention}, have been rapidly adopted in neural networks handling text \cite{brown2020language,devlin2018bert,liu2019roberta,radford2019language,vaswani2017attention}, speech \cite{karita2019comparative,luscher2019transformers}, and even for more complex tasks such as  function integration or solving differential equation~\cite{lample2019deep}. In computer vision, DeTR~\cite{carion2020end} and Vision Transformers~\cite{dosovitskiy2020image} (ViT) have deeply influenced the design of architectures in a short period of time. Most of the architectures introduced since ViT can be regarded as some form of hybridisation of transformers with convolutional neural networks, as illustrated by the hierarchical transformers~\cite{fan2021multiscale,graham2021levit,liu2021swin,Wang2021PyramidVT}, or conversely by convolutional neural networks with design elements inspired from ViT~\cite{liu2022convnet,touvron2021augmenting}, or even multi-layer perceptrons adopting designs inspired by transformers~\cite{melaskyriazi2021doyoueven,liu2021mlp,tolstikhin2021MLPMixer,Touvron2021ResMLPFN,ding2021repmlp}. 

In our case we build upon the basic ViT design of Dosovitskiy. 
Its design is governed by a small hyper-parameter space, and as such is less engineered than some  recent follow-up architectures. 
With a proper training procedure~\cite{Steiner2021HowTT,Touvron2020TrainingDI,wightman2021resnet}, it achieves interesting performance/complexity trade-offs. 
It is also versatile: it can be effectively combined with hierarchical  detection or segmentation frameworks~\cite{el2021xcit}. 
Importantly, in spite of limited built-in priors, it has demonstrated great potential when combined with self-supervised learning, either with contrastive methods~\cite{caron2021emerging,chen2021empirical} or for reconstruction-based techniques like BeiT~\cite{bao2021beit}  or other forms of masked auto-encoders~\cite{el2021large,he2021masked,wei2021masked,xie2021simmim,dong2021peco,zhou2021ibot}. 

\subsection{Experimental setting}

\mypar{ViT models.}
We consider the vanilla ViT models initially introduced by Dosovitskiy et al.~\cite{dosovitskiy2020image} as well as the smaller ones proposed by Touvron et al.~\cite{Touvron2020TrainingDI}. Therefore we use the initial pooling method that is based on a so-called class token.  
We only consider transformers operating on 16$\times$16 patches. Decreasing this patch size improves the results but significantly increases the model complexity.  

\mypar{Training procedure.}
To prevent overfitting, we adopt an existing training setting, namely the A2 procedure  of Wightman et al.~\cite{wightman2021resnet}. 
It uses a binary cross entropy loss and fixes the setting of most of the hyper-parameters.%
Wightman et al.'s A2 procedure was originally designed for training ResNet-50 models, and  requires a few modifications when adopting it for ViTs to get strong  performance and ensure  sufficient stability:
\begin{itemize}
\item \emph{The learning rate} should be reduced compared to ResNet-50. We set it to $lr=4.10^{-3}$ for ViT-Ti and ViT-S and to $lr=3.10^{-3}$ for ViT-B and ViT-L. 
\item \emph{Stochastic depth drop-rate $sd$:} we adjust it per model following Touvron et al.~\cite{touvron2021going}. It is not used for ViT-Ti. We fix $sd=0.05$ for Vit-S, $sd=0.1$ for ViT-B and $sd=0.4$ for Vit-L. 
\end{itemize}
We observe that LayerScale~\cite{touvron2021going} significantly improves the performance when training large models, and that in that case a longer training is also beneficial. 
Therefore in addition to our main baseline where we train during 300 epochs without LayerScale, like in DeiT and in the A2 procedure of Wightman et al.~\cite{wightman2021resnet}, we consider another one that is trained for 400 epochs with LayerScale (LS).  

\mypar{Evaluation.}
Unless specified otherwise, we train our models on the ImageNet-1k dataset~\cite{Russakovsky2015ImageNet12},  and evaluate the top-1 accuracy on its validation set.
All experiments are carried with seed 0. 
Since we have adjusted a low number of hyper-parameters, and since we share them across models except stochastic depth, we do not expect much overfitting. 
Nevertheless we also evaluate our models with the same metric on ImageNet-V2~\cite{Recht2019ImageNetv2} (matched frequency), which provides a separate test set, to provide a complementary view on the results.

\begin{table}[t]
    \caption{Baseline models and their performance on ImageNet1k-val top1 accuracy at resolution 224$\times$224. We adopt common models with their default parametrization: Vit-B and Vit-L~\cite{dosovitskiy2020image} and Vit-Ti and ViT-S~\cite{Touvron2020TrainingDI}, all with patch size of 16$\times$16.  
    Baseline results trained with LayerScale or not (LS)~\cite{touvron2021going}, and for 300 or 400 epochs of training. 
    \label{tab:base_archi}}
\centering
\vspace{-1.0em}
{
\setlength{\tabcolsep}{3pt} 
\begin{tabular}{l|rrr|rrr|cc|cc}
\toprule
        & &&&  params & Flops & speed & \multicolumn{2}{c}{300 epochs} & \multicolumn{2}{|c}{400 ep.+LS} \\
Model   & depth  & width  & heads & ($\times 10^6$) &  ($\times 10^9$) & (im/s) & val & v2 & val & v2 \\
\midrule
ViT-Ti/16  & 12\ \  & \pzo192 & \pzo3 &  5.7  & 1.3    & 3796  & 72.7 & 60.3   &  73.5   & 61.4  \\ 
ViT-S/16   & 12\ \  & \pzo384 & \pzo6 &  22.1  & 4.6   & 1827  & 79.7 & 68.5   &  80.7   & 69.3  \\ 
ViT-B/16   & 12\ \  & \pzo768 & 12    &  86.6   & 17.6 & 799   & 82.2 & 71.2   &  82.7   & 72.2  \\ 
ViT-L/16   & 24\ \  & 1024    & 16    & 304.4  & 61.6  & 277   & 83.0 & 72.4   &  84.0   & 73.7  \\ 
\bottomrule
\end{tabular}
}\end{table}

\subsection{Baselines} 

We report the results of our baseline in Table~\ref{tab:base_archi}.
With the few adaptations that we have done, our training procedure outperforms existing ones for supervised training for the model sizes that we consider, see Appendix~\ref{sec:baseline_supmat} (Table~\ref{tab:comp_1k_method}). 
Note that all our models use a patch size of 16$\times$16 as in Dosovitskiy et al.~\cite{dosovitskiy2020image}. 
Unless specified, our experiments are carried out with images of size 224$\times$224. 

\section{Depth vs Width: Parallel ViT}
\label{sec:seq_vs_par}

A recurrent debate in neural architecture design is on how to balance width versus depth. The first successful neural networks on Imagenet~\cite{Krizhevsky2012AlexNet,Simonyan2015VGG} were not very deep, for instance the 22-layer GoogleNet~\cite{Szegedy2015Goingdeeperwithconvolutions} was regarded as deep in 2014's standards. 
This has changed with ResNets~\cite{He2016ResNet,He2016IdentityMappings}, for which going deeper was hindering significantly less the optimization due to the residual connections. After its introduction, some researchers have investigated alternative choices for trading depth against width~\cite{Zagoruyko2016WideRN,Huang2016DeepNW,Ding2021RepVGGMV}, like Wide Residual Networks~\cite{Zagoruyko2016WideRN}. 

Recently, there has been a renewed interest for wider architectures with attention~\cite{Goyal2021NondeepN,Li2019SelectiveKN}. For instance the Non-deep Networks~\cite{Goyal2021NondeepN} proposes an architecture with several parallel branches whose design is more complex. 
In our work, we aim at proposing a much simpler and flexible alternative that builds upon a regular ViT in a more straightforward manner.

\subsection{Preliminary discussion on width versus depth for ViT} 

The ViT architecture of Dosovitskiy et al.~\cite{dosovitskiy2020image} is parametrized by three  quantities: the width (i.e., the working dimensionality $d$), the depth, and the number of heads. We do not discuss the latter. Increasing depth or width increases the capacity of the model and usually its accuracy. For the most common ViT models that we report in Table~\ref{tab:base_archi}~\cite{dosovitskiy2020image,Touvron2020TrainingDI}, width and height are scaled together. 
Below, we discuss the different pros and cons for favoring width versus depth.

\mypar{Parametrization \& Optimization.}
The compositionality of the layers is better with deeper networks. This was one of the decisive advantage of ResNet once optimization issues were  solved by residual connections. Yet too much depth hinders optimization, even with residual connections. Some solutions have been proposed to address this issue for ViTs~\cite{touvron2021going}, showing that transformers benefit from depth when trained with improved optimization procedure.

\mypar{Separability.} %
In image classification, the spatial features are ultimately  projected~\cite{Krizhevsky2012AlexNet} or pooled~\cite{He2016ResNet} into a %
high-dimensional latent vector that is subsequently fed to a linear classifier. 
The dimensionality of this vector %
should be high enough so that the classes are linearly separable. Hence it is typically larger for tasks involving many classes. For instance in ResNet-50 it has dimension 512 when applied to CIFAR, but 2048 for ImageNet. 
In ViT, the width is identical to the working dimensionality of each patch, and is typically smaller than with ResNet, possibly limiting the separation capabilities. 
Besides, a larger dimension of the latent vector tend to favor overfitting. In this regard the compromise between capacity and overfitting is subtle and depends size of the training set~\cite{Steiner2021HowTT}. 

\mypar{Complexity.} In ViT, the different complexity measures are affected differently by width and depth. 
Ignoring the  patch pre-processing and final classification layer, which contribute to complexity in a negligible manner, then we have:
\begin{itemize}
    \item \emph{The number of parameters} is proportional to depth and a quadratic function of the width. 
    \item \emph{The compute,} as determined by FLOPS, is similarly proportional to the depth and quadratic in width. 
    \item \emph{The peak memory usage at inference time} is constant when increasing the depth for a fixed width, but it is quadratic as a function of width.
    \item \emph{The latency} of wide architectures is in theory  better as they are more parallel, but  actual speedups depend on implementation and hardware.
\end{itemize}

\subsection{Parallelizing ViT } 

\def \mhsa  {\mathrm{mhsa}} 
\def \ffn   {\mathrm{ffn}} 

We propose and analyze flattening vision transformers by grouping layers following the scheme presented in the introduction. Let us consider a sequence of transformer blocks defined by the functions $\mhsa_l(\cdot)$, $\ffn_l(\cdot)$, $\mhsa_{l+1}(\cdot)$ and $\ffn_{l+1}(\cdot)$. Instead of sequentially processing the input $x_l$ in four steps as done in the usual implementation:
\begin{align}
    x'_{l+1} & = x_l + \mhsa_l(x_l),              & x_{l+1}  = &\  x'_{l+1} + \ffn_l(x'_{l+1}),\nonumber \\
    x'_{l+2} & = x_{l+1} + \mhsa_{l+1}(x_{l+1}),  & x_{l+2}  = &\  x'_{l+2} + \ffn_{l+1}(x'_{l+2}), 
\label{eq:normal}
\end{align}
we replace this composition by two parallel operations:
\begin{align} 
    x_{l+1} & = x_l+ \mhsa_{l,1}(x_l) + \mhsa_{l,2}(x_l), \nonumber \\
    x_{l+2}  & =x_{l+1} + \ffn_{l,1}(x_{l+1})  + \ffn_{l,2}(x_{l+1}).  
\label{eq:approx}
\end{align} 
This reduces the number of layers by two for a given number of MHSA and FFN blocks. Conversely, there is twice the amount of processing in parallel. 
The intuition behind this parallelization is as follows: as networks get deeper, the contribution of any residual block $r(\cdot)$, be it $\mhsa(\cdot)$ or $\ffn(\cdot)$, becomes increasingly smaller with respect to the overall function. Therefore, the approximation  $\forall r,r' \;\;  r'(x+r(x)) \approx  r'(x)$ becomes increasingly satisfactory, and it is easy to check that if this approximation is true, eq.~(\ref{eq:normal}) and (\ref{eq:approx}) are equivalent. 

Our strategy is different from taking transformers with a larger working dimensionality, which leads to different trade-offs between accuracy, parameters, memory and FLOPS, as discussed in our experiments. 
In contrast to increasing the working dimension, which increases the complexity quadratically as discussed above, 
our modification is neutral with respect to parameter and compute. 

Depending on whether we effectively parallelize the processing, the peak memory usage at inference time and the latency are modified. 
Note that rather than just two, we can choose to process any number of blocks in parallel; falling back to the sequential design if we process a single block in each layer.

\subsection{Experiments} 

\noindent \textbf{Notation.} We adopt the standard naming convention of previous work~\cite{dosovitskiy2020image,Touvron2020TrainingDI}  to use the postfixes Ti/S/B/L to identify the working dimensionality of the models, i.e., the column ``width'' in Table~\ref{tab:base_archi}. We append the depth $N$ to indicate variations on the number of pairs of layers (MHSA,FFN)~\cite{touvron2021going}. 
For instance, ViT-B24  has the same width as a ViT-B12 but with twice the depth, i.e., 24 pairs of MHSA and FFN layers instead of 12. 
For our parallel models, we specify both the depth and the number of parallel branches: ViT-B12$\times$2  has twice the number of residual modules as a ViT-B12. It includes a total of 12$\times$2=24 pairs of MHSA and FFN layers. Therefore it has the same complexity as the ViT-B24 model (a.k.a.\ ViT-B24$\times$1).

\begin{figure}[p]
\begin{minipage}{0.48\linewidth}
\centering
\includegraphics[height=0.65\linewidth]{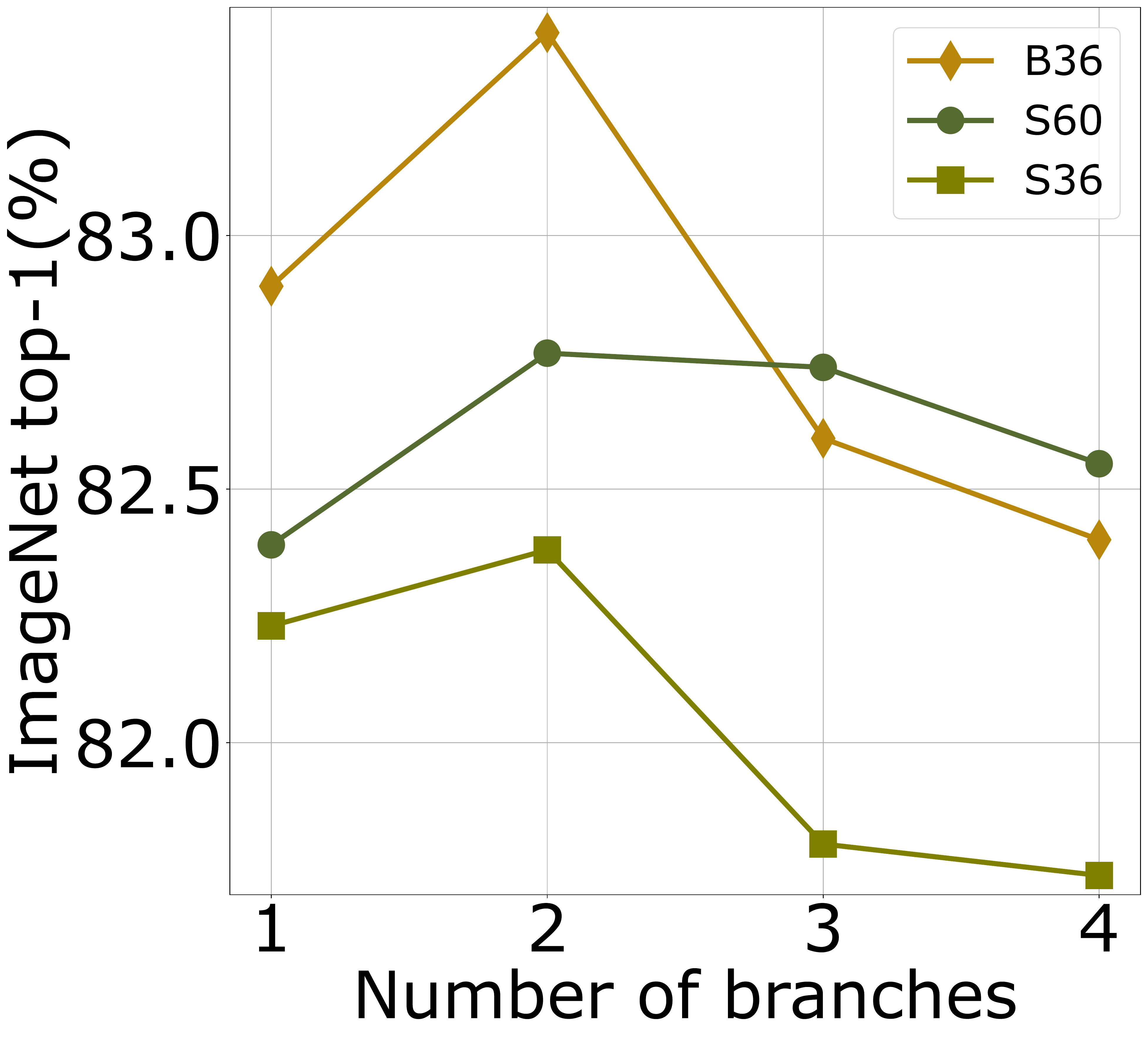}
    \caption{Impact of the parallelism on performance for a given model size (ViT-S36, -S60 and -B36) and 1--4 parallel branches.
    \label{fig:seq_vs_par_nbbranches}}
\end{minipage}
\hfill
\begin{minipage}{0.48\linewidth}
\centering
 \setcounter{figure}{1} 
    \renewcommand{\figurename}{Table}
    \caption{Impact of the training on parallel and sequential models.     
    \label{tab:seq_vs_par_optim}}
        \smallskip
    \scalebox{0.85}{
    \begin{tabular}{lcc@{ \ }|cc}
    \toprule
                     & Number of &  &  \multicolumn{2}{c}{ImNet top1} \\
        Model        & Epochs &  LS &   -val  & -v2\\
        \midrule
                     &  300   & \xmark  &   82.9  &  72.2        \\
         sequential: &  300   & \cmark  &   83.9  &  73.2         \\
         ViT-B36x1   &  400   & \xmark  &   83.4  &  72.5        \\
                     &  400   & \cmark  &   84.1  &  73.9        \\
        \midrule
                     &  300   & \xmark  &   83.3  &  72.4        \\
        parallel:    &  300   & \cmark  &   83.8  &  73.3        \\
        ViT-B18x2    &  400   & \xmark  &   83.4  &  73.1        \\
                     &  400   & \cmark  &   84.1  &  73.5        \\
    \bottomrule
    \end{tabular}}
\end{minipage}

\begin{minipage}{0.48\linewidth}
\centering
\includegraphics[height=0.65\linewidth]{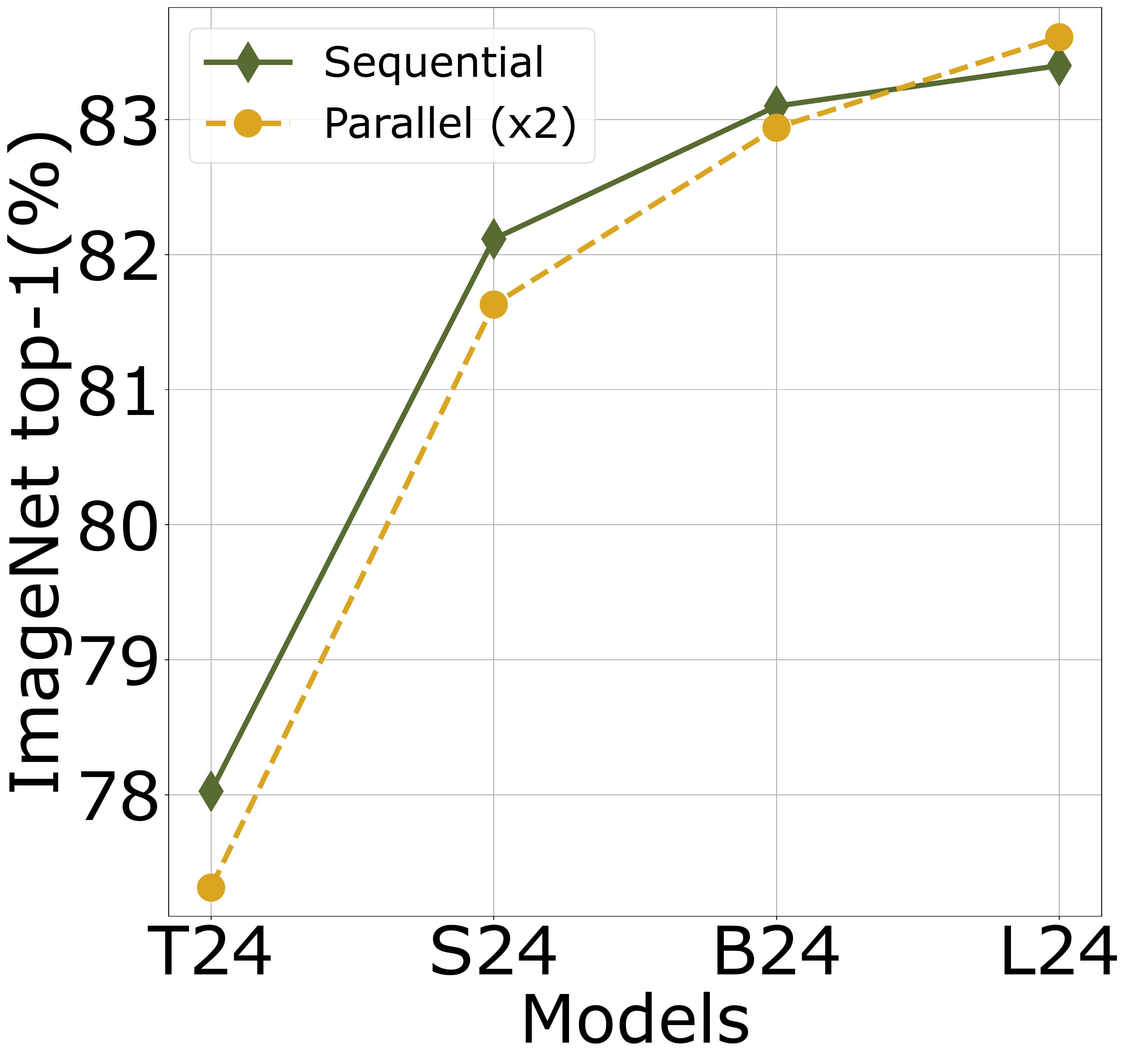}
    \setcounter{figure}{1} 
    \caption{Impact of model width (T:192, S:384, B:768, L:1024). We train the two L24 with LS to avoid optimization issues. 
    \label{fig:seq_vs_par_models}}
\end{minipage}
\hfill
\begin{minipage}{0.48\linewidth}
\centering
        \setcounter{figure}{2} 
        \renewcommand{\figurename}{Table}
    \caption{Comparison of parallel models with more blocks with models with a higher working dimensionality. L24$\times$1, B36$\times$1 and B18$\times$2 trained with LS.  
    \label{tab:par_vs_bigger}}
    \smallskip
    \scalebox{0.75}{
    \begin{tabular}{rccc|ll}
    \toprule
                 &  \#params        &  Flops        & Mem.&  \multicolumn{2}{c}{ImNet top1} \\
        Model    &  ($\times$$10^6$) & ($\times$$10^9$) &(MB)    &  \ \ \ -val & \ \ \ -v2   \\
        \midrule
        B12x1    &   86.6   &  17.6   &   2077      & 82.2\stdminus{0.06}  &  71.0\stdminus{0.26}        \\
        S48x1    &   85.9     &   18.3     &   1361       &82.3  & 72.0 \\
        S24x2    &   85.9      &   18.3     &   1433       &82.6  & 72.3 \\
        \midrule                 
        L24x1    & 304.4   &  61.6   &   3788        &83.4  &  73.3 \\
        B36x1    & 256.7   &  52.5      &   3071        &83.9  & 73.2  \\
        B18x2    & 256.7   &  52.5      &   3217        &83.8  & 73.3  \\
    \bottomrule
    \end{tabular}}
\end{minipage}

\newcommand{\ccg}{\cellcolor{Goldenrod!75}}

\begin{minipage}{0.48\linewidth}
\centering
\includegraphics[height=0.65\linewidth]{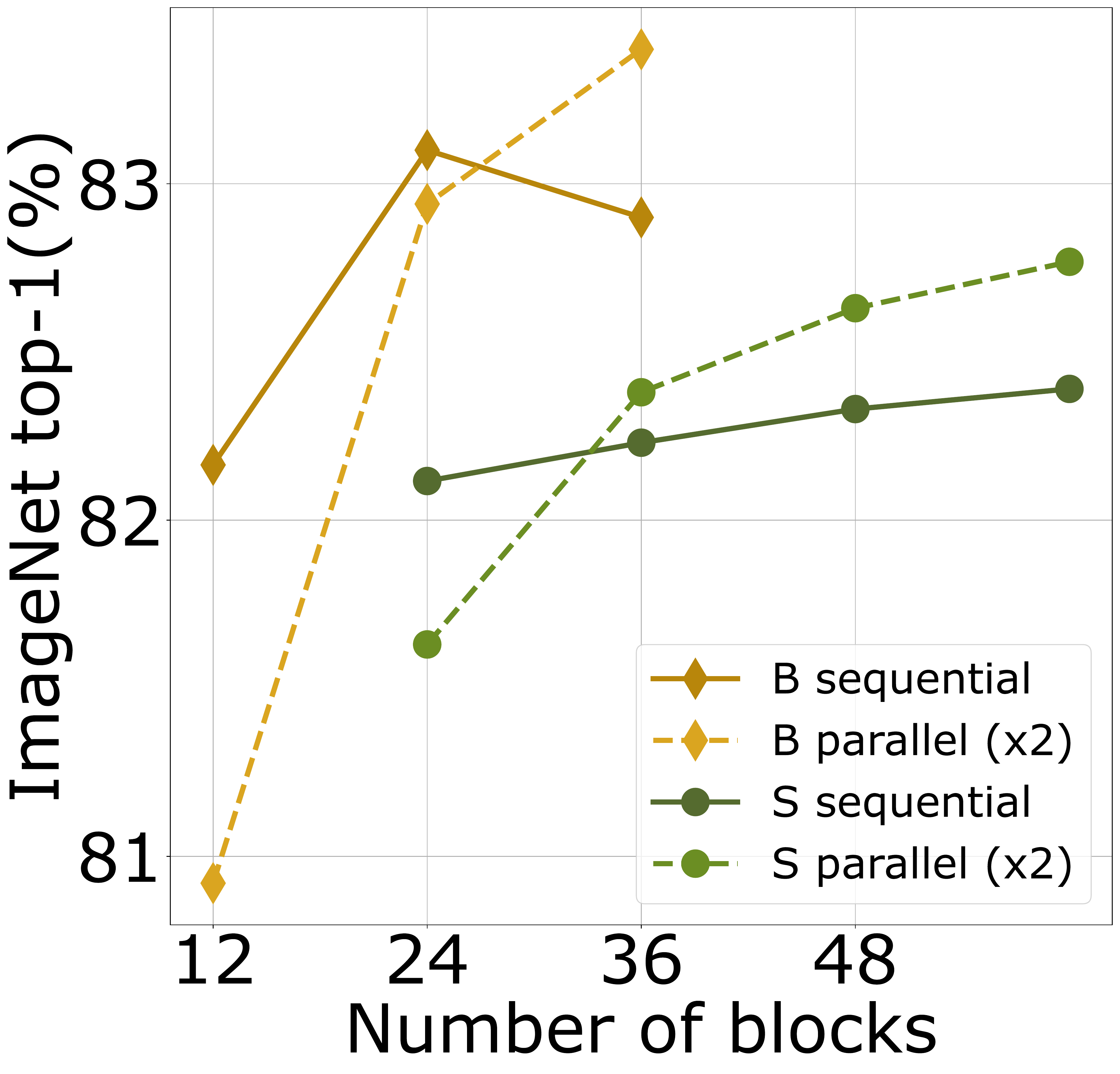}
    \setcounter{figure}{2} 
    \caption{Sequential vs.\  parallel ViT-S and -B when varying the number of blocks. 
    \label{fig:eq_vs_par_total_blocks}}
\end{minipage}
\hfill
\begin{minipage}{0.48\linewidth}
        \setcounter{figure}{3} 
        \renewcommand{\figurename}{Table}
    \caption{Throughput for ViT-S18$\times$2 and ViT-B18$\times$2 (im/s). With parallel ViT, the residual blocks can be processed either sequentially (seq) or in parallel (par). 
    \label{tab:seq_vs_par_latency}}
    \setlength{\tabcolsep}{2.5pt} 
    \scalebox{0.85}{
    \begin{tabular}{c|rrrr|rrrr}
    \toprule
      batch  & \multicolumn{4}{c}{ViT-S18x2} &  \multicolumn{4}{c}{ViT-B18x2} \\
      \cmidrule(l){2-5} \cmidrule(l){6-9} 
      size   &  seq  & par & best & gain   & seq  & par & best & gain   \\  
      \midrule
        1	&   44	 &  \ccg61	&  \ccg61	 &  38\%	&  42	& \ccg61	&  \ccg61	& 45\%  \\
        2	&   84	 & \ccg123	& \ccg123	 &  46\%	&  80	& \ccg117	& \ccg117	& 47\%  \\
        4	&  168	 & \ccg245	& \ccg245	 &  46\%	& 155	& \ccg187	& \ccg187	& 21\%  \\
        8	&  334   & \ccg474	& \ccg474	 &  42\%	& \ccg230	& 211	& \ccg230	& 0\%  \\
        16	&  \ccg569   & 518	& \ccg569	 &  0\%	    & \ccg266	& 231	& \ccg266	& 0\%  \\
        32	&  \ccg616   & 556	& \ccg616	 &  0\%	    & \ccg276	& 245	& \ccg276	& 0\%  \\
        64	&  \ccg647   & 575  & \ccg647	 &  0\%  	& \ccg286	& 248	& \ccg286	& 0\%  \\
    \bottomrule
    \end{tabular}}
\end{minipage}

\end{figure}

\mypar{Comparison of sequential and parallel ViTs.} 
In Figure~\ref{fig:seq_vs_par_nbbranches}, we compare the performance of  sequential and parallel models of a fixed complexity. We fix the total number of blocks, i.e.\ pairs of MHSA and FFN layers, which determines the number of parameters and FLOPS, and we consider different possible of branches that leads to the same total number of blocks. 
For instance 36 can be obtained as the sequential ViT 36$\times$1, or the parallel ViTs 18$\times$2, 12$\times$3 or 9$\times$4. %

We observe that, amongst the parallel and sequential models, the best performance is obtained with two parallel branches for all tested model capacities.
The performance is comparable between the S20$\times$3 and S30$\times$2 for ViT-S60, but generally using more than two parallel branches is not  favorable in terms of accuracy and we do not discuss them further. 
Note that Figure~\ref{fig:seq_vs_par_nbbranches} compares ViT models with a relatively large number of blocks (36 and 60). 
This is the case where sequential models are relatively difficult to optimize due to their depth. The parallel models with  two branches are easier to train, while being deep enough to benefit from layer compositionality.

In Figure~\ref{fig:seq_vs_par_models}, we consider models with only 24 pairs (MHSA,FFN) and a varying width. Here we observe that the smallest models ViT-Ti and ViT-S are better in their sequential version. This is because are easy to optimize up to 24 layers.  
The B24$\times$1 and B12$\times$2 achieve  comparable performance. In contrast, the ViT-L12$\times$2 is stronger than its sequential counterpart, which is more difficult to optimize
even though we  used LS for this size; without LS its performance is  
83\% at 300 epochs. 

In Figure~\ref{fig:eq_vs_par_total_blocks}, we compare the performance of sequential and parallel as a function of the number of blocks for ViT-S and ViT-B. Our observations concur with our previous findings: the parallel version is 
more helpful for the deeper and higher capacity models that  are more difficult to optimize; our parallelization scheme alleviates this issue. 

\mypar{Impact of optimization.} 
In Table~\ref{tab:seq_vs_par_optim}, we provide results with LayerScale~\cite{touvron2021going}, which helps the optimization of the biggest models. It improves the performance of both sequential and parallel models, which end up  approximately on par. Hence, for models big enough and with proper optimization,  sequential and parallel ViTs are roughly equivalent. 

\mypar{Increasing the number of modules or the working dimensionality?} 
Table~\ref{tab:par_vs_bigger} provides a comparison between different ViT architectures: sequential, parallel,  and with larger working dimensionality. 
We approximately adjust the complexity in terms of parameters and FLOPS, yet this means that  ViT models with  larger working dimensionality have a higher peak memory usage with typical implementation.  
In both tested settings the sequential and parallel models yield substantially higher accuracy than the models with larger working dimensionality. The sequential and parallel models are comparable with 36 blocks. The parallel model is better in the case of 48 blocks due to the increased depth of the sequential model.

\mypar{Latency.}   On a commodity V100 GPUs, we observe a significant speed-up in the case of per-sample processing, with also some gains for small batch sizes with relatively small models, see Table~\ref{tab:seq_vs_par_latency}. 
This comparison is based on a simple implementation of our parallel architecture,  which is  suboptimal due to the  lack of a specific CUDA kernel. 
Overall our measurements  suggest  specific hardware or  kernels are required to obtain compelling benefits in terms of throughput.

\section{Fine-tuning attention is all you need}
\label{sec:finetune}

In this section we focus on fine-tuning ViT models,  either to adapt the the model to larger image resolutions or to address different downstream classification tasks.
In particular, we consider an approach where we only fine-tune the weights corresponding to the MHSA layer, see Figure~\ref{fig:attn_finetuning}.
We analyse the impact in terms of prediction accuracy and  savings in  processing complexity, peak memory usage and parameter count. 
As we will see, our choice is significantly better than alternative ones, such as fine-tuning the parameter-heavy FFN layers. 

It is  common to train networks at lower resolution and fine-tuning it at a higher target resolution. This saves a significant amount of compute at training time, and typically also improves the  accuracy of the network at the target resolution%
~\cite{Touvron2019FixRes}. 
This is because it reduces the discrepancy between the scale of the images seen at train and at test time that is induced by common data augmentation. 
Fine-tuning is also the paradigm associated with foundation models in general and to the concept of transfer learning itself~\cite{Yosinski2014HowTA,oquab2014learning,Guo2019SpotTuneTL}. 
A recent line of work explores adaptation of pre-trained models with various types of adapter modules with a small amount of task-specific parameters~\cite{berriel19iccv,Houlsby2019ParameterEfficientTL,mancini18eccvw,Rebuffi2018EfficientPO,Mahabadi2021ParameterefficientMF,Pfeiffer2020AdapterHubAF}. 
In our work, instead, we focus on fine-tuning vanilla ViTs.

\begin{figure}[t]
\begin{minipage}{0.55\linewidth}
\includegraphics[width=0.9\linewidth]{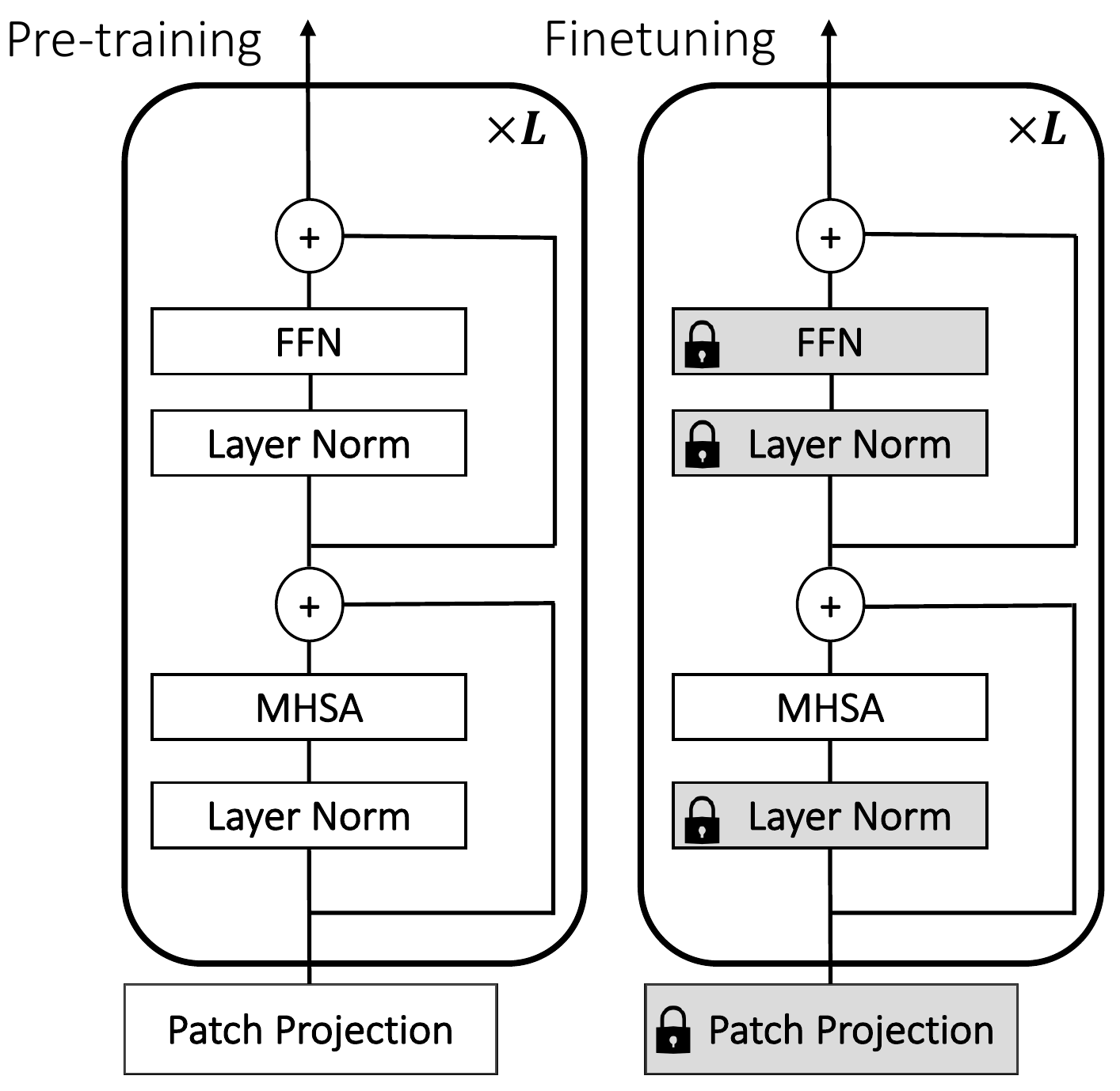}
\end{minipage}%
\hfill
\begin{minipage}{0.45\linewidth}
\hfill
\begin{minipage}{0.9\linewidth}
\includegraphics[trim={1.0cm, 0.5cm, 1cm, 0.5cm}, width=0.9\linewidth]{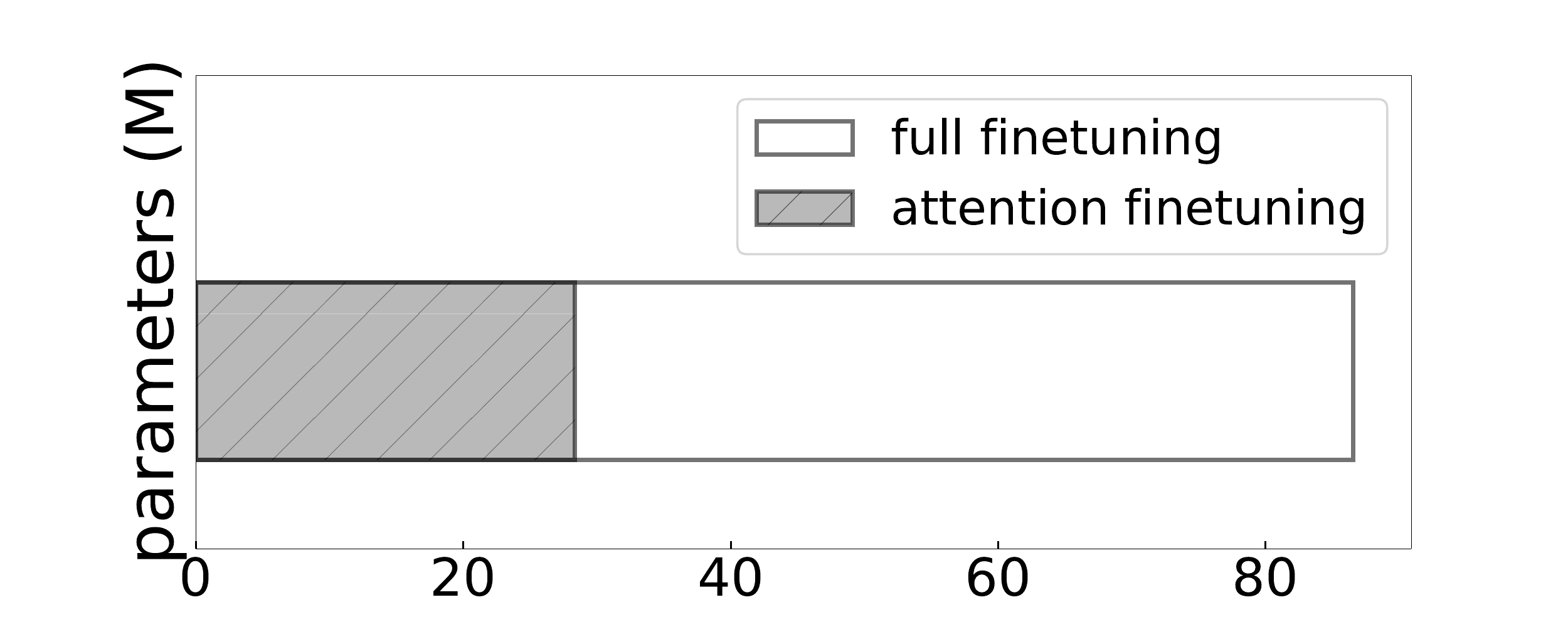}
\medskip
\includegraphics[trim={1.0cm, 0.5cm, 1cm, 0.5cm}, width=0.9\linewidth]{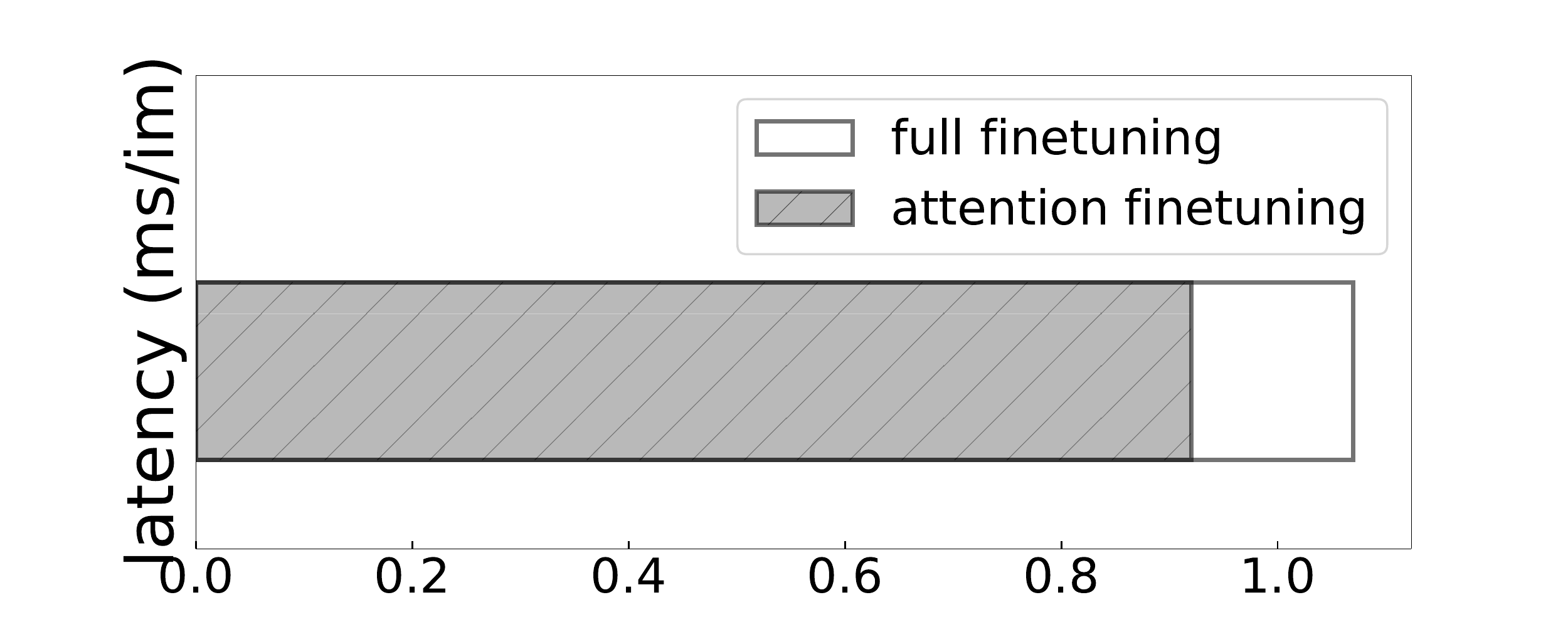}
\medskip
\includegraphics[trim={1.0cm, 0.5cm, 1cm, 0.5cm}, width=0.9\linewidth]{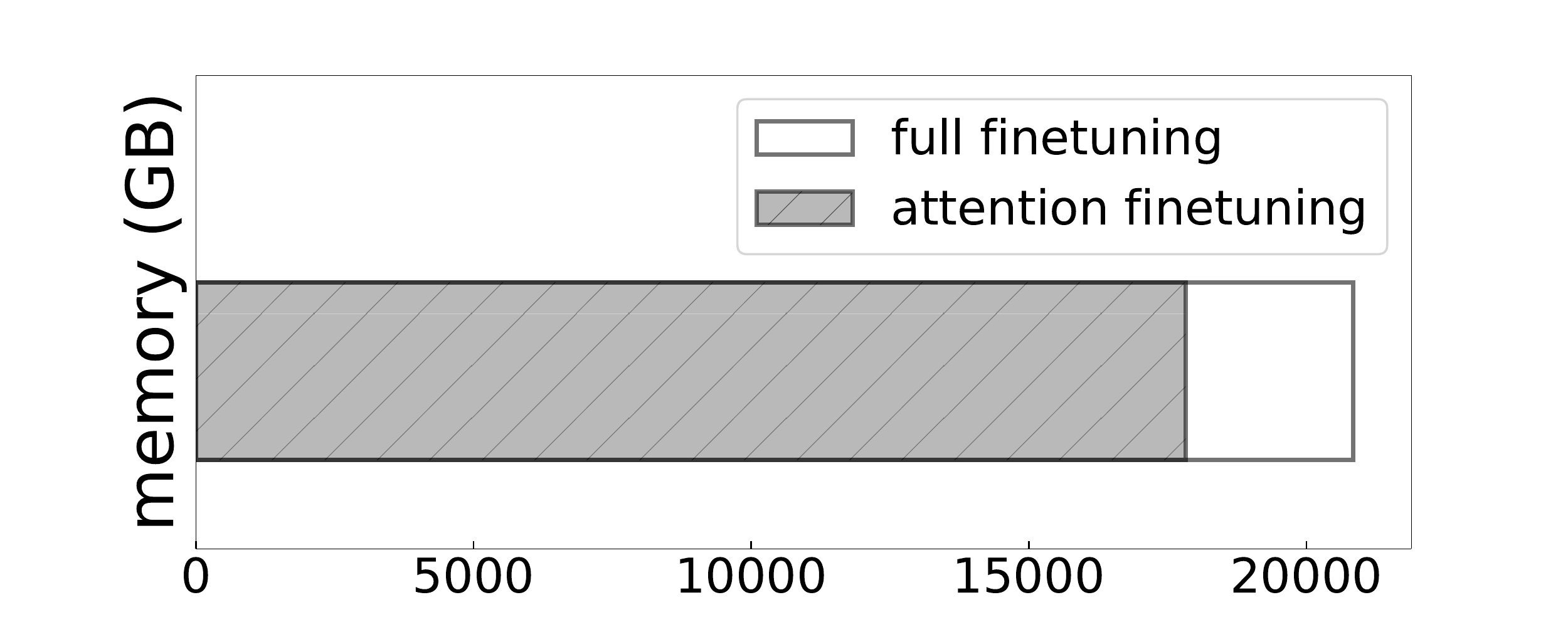}
\end{minipage}

\end{minipage}%
    \setcounter{figure}{3} 
    \caption{Fine-tuning  the weights of the self-attention layer only (middle panel) leads to  savings during  fine-tuning in peak memory usage and computational cost. 
    It also leads to important savings in the number of parameters when a model is fine-tuned for  multiple resolutions or  multiple downstream classification tasks. 
    }
    \label{fig:attn_finetuning}
\end{figure}

\mypar{Fine-tuning at different resolutions.} 
In Table~\ref{tab:attn_finetuning}, we report results with fine-tuning ViT-S, ViT-B and ViT-L at 384$\times$384 resolution for models pre-trained at 224$\times$224. 
Solely fine-tuning the MHSA weights provides results that are within standard deviation ($\pm{0.1}$) from a full fine-tuning both on ImageNet-val and ImageNet-V2. 
This is not the case when fine-tuning the FFN layers, while these contain twice the number of parameters of MHSA.  
Note, our pre-trained models have been  trained long enough (400 epochs) to ensure convergence. 

There are only advantages to use this approach when fine-tuning at higher resolution as opposed to doing a full fine-tuning, as we get substantial savings  in terms of parameters, latency, and peak memory usage for free, see  Figure~\ref{fig:attn_finetuning} (right panels). 
First, the fine-tuning stage requires 10\% less memory on the GPU, which is especially interesting in the context of high-resolution fine-tuning where the higher images require more memory. 
The training is also 10\% faster, as less gradients are computed. 
Finally, the attention weights correspond to approximately one third of the weights. Therefore,
if one wants to use multiple models fine-tuned for  different input resolutions, we save 66\% of the storage for each additional model. 

\def \mysp {\hspace{8pt}}
\begin{table}
      \setcounter{table}{4} 
    \caption{Comparison of full finetuning of all weight (full), finetuning of the MHSA layer weights only (attn) and of the FFN layer only (ffn) when adapting  models at resolution $384\times384$ on ImageNet-1k from model pre-trained at $224\times224$. We compare finetuning with the SGD and AdamW~\cite{Loshchilov2017AdamW} optimizers.  
    \label{tab:attn_finetuning}}

    \hfill
    \scalebox{0.75}{
    \begin{tabular}{c@{\mysp}|@{\mysp}ccc@{\mysp}|@{\mysp}ccc}
        \multicolumn{7}{c}{\bf ImageNet1k-val top1 acc.  } \\
        \toprule
        \multirow{2}{*}{Model} & \multicolumn{3}{@{\ }l@{\ }}{AdamW$\uparrow$384} & \multicolumn{3}{c}{SGD$\uparrow$384} \\
         & full & attn & ffn & full & attn & ffn \\
         \midrule
         ViT-S & \textbf{82.7} &	82.5 &	82.2&	82.6 &	82.3 &	82.0\\
         ViT-B & \textbf{84.3} &	\textbf{84.3} &	84.1 &	84.3&	84.2 &	84.0\\
         ViT-L & \textbf{85.5} & \textbf{85.5} & 85.2	 & 85.4	 & 85.3	& 85.1\\
         \bottomrule
    \end{tabular}}
    \hfill
    \scalebox{0.75}{
    \begin{tabular}{c@{\mysp}|@{\mysp}ccc@{\mysp}|@{\mysp}ccc}
        \multicolumn{7}{c}{\bf ImageNet1k-V2 top1 acc.  } \\
        \toprule
        \multirow{2}{*}{Model} & \multicolumn{3}{@{\ }l@{\ }}{AdamW$\uparrow$384} & \multicolumn{3}{c}{SGD$\uparrow$384} \\
         & full & attn & ffn & full & attn & ffn \\
         \midrule
         ViT-S &72.5 & 72.4 & 71.6  & 72.5 & 72.2 & 71.1 \\ 
         ViT-B & 73.7 & 74.0  & 73.6 & 74.0 & 73.9 & 73.7 \\ 
         ViT-L & 75.5 & 75.4 & 75.2 & 75.6 & 75.1 &  75.0 \\ 
         \bottomrule
    \end{tabular}}
\hfill  \smallskip
    
\end{table}

\mypar{Fine-tuning on different datasets.} 
We now evaluate  our approach when transferring  ViTs pre-trained on ImageNet to  different  downstream classification tasks by fine-tuning. 
We consider public benchmarks whose characteristics and references are given in Appendix~\ref{sec:dataset_supmat}.

In Table~\ref{tab:finetuning_attn_TL} we report the performance for different fine-tuning strategies. Here we make different observations. 
First, for the \textbf{smallest datasets},  namely CARS and Flower, fine-tuning only the MHSA layers is an excellent strategy. It is even better than full-tuning. Our interpretation is that restricting the number of weights has a regularizing effect. 
The conclusion is more mixed with the \textbf{largest datasets}, in particular iNaturalist, where we observe a significant gap between the full fine-tuning and our solution for the ViT-S.  This could be expected: in this case there are more images to learn from and new classes that were not seen before the fine-tuning stage. 
Restricting the fine-tuning to MHSA layer allows modifying only a relatively small number of parameters. FFN layers have twice more weights and leads to better results in that case.  
This limitation tends to disappear with the \textbf{larger ViT-L models}, for which the the capacity of the MHSA is much larger and therefore sufficient. Our strategy is therefore interesting in the typical use-cases of foundation models, which are very large models that are fine-tuned on a variety of downstream tasks.

\begin{table}
       \caption{Transfer learning experiments: we compare full finetuning, finetuning of attention only and finetuning with ffn only on 6 transfer learning dataset with 3 differents ViT models pre-trained on ImageNet-1k only. %
    \label{tab:finetuning_attn_TL}}
    \centering
    \scalebox{0.9}{
    \begin{tabular}{c|ccc|cccccc}
        \toprule
        \multirow{2}{*}{Model}  & \multicolumn{3}{c|}{Finetuning}  & \multirow{2}{*}{INAT-18} & \multirow{2}{*}{INAT-19} & \multirow{2}{*}{CIFAR-10} & \multirow{2}{*}{CIFAR-100} & \multirow{2}{*}{CARS} & \multirow{2}{*}{Flowers}\\
        \cmidrule{2-4}
         &  full & attn & ffn \\
         \midrule
         \multirow{3}{*}{ViT-S} 
         & \cmark & \xmarkg & \xmarkg & \textbf{68.0} & \textbf{73.9} & \textbf{98.9} & \textbf{90.5} & 89.7 & 96.8\\
         & \xmarkg & \cmark & \xmarkg & 60.6 & 68.7 & 98.7 & 89.1 & \textbf{89.8} & \textbf{96.9} \\
         & \xmarkg & \xmarkg & \cmark & 64.4 & 72.5 & 98.9 & 90.1 & 88.3 & 96.1 \\
         \midrule
         \multirow{3}{*}{ViT-B} &
         \cmark & \xmarkg & \xmarkg & \textbf{74.1} &	\textbf{78.2} &	\textbf{99.3} &	\textbf{92.5} &	92.7 &	97.8\\
         & \xmarkg & \cmark & \xmarkg & 71.1 &	75.7 &	99.2 &	91.8 &	\textbf{92.8}& 	\textbf{98.5}\\
         & \xmarkg & \xmarkg & \cmark & 73.3 & 77.3 & \textbf{99.3} & 92.1 & 88.9 & 97.5 \\
         \midrule    
          \multirow{3}{*}{ViT-L} 
         & \cmark & \xmarkg & \xmarkg & \textbf{75.9}	& \textbf{79.7} &	\textbf{99.3}	& \textbf{93.2}	& \textbf{93.8}	& 98.3\\
         & \xmarkg & \cmark & \xmarkg & 75.3 &	78.7 & 99.2 & 92.7 & \textbf{93.8} & \textbf{98.4}\\
         & \xmarkg & \xmarkg & \cmark & 75.4 & 79.3 & 99.2 & 93.0 & 93.0 & 97.6\\
         \bottomrule   
    \end{tabular}}
\end{table}

\section{\makebox{Patch preprocessing for Bert-like self-supervised learning}}
\label{sec:stem}

The  original ViT paper~\cite{dosovitskiy2020image}  considered to include convolution instead of patch projection in the network design. Several recent papers~\cite{Wu2021CvTIC,Han2021TransformerIT,Yuan2021TokenstoTokenVT,Wang2021PyramidVT,graham2021levit,Xiao2021EarlyCH} advocate this choice to  include  a small pre-processing network in the architecture, instead of a simple patch projection. Most of the pre-processing subnetworks that have been considered are based on convolutions, and are often referred to as ``convolutional stems''. Small transformers have also been considered~\cite{Yuan2021TokenstoTokenVT}. 

While these patch pre-processing designs have been developed to improve accuracy and/or  stability, there are some remaining questions regarding their design and flexibility. 
First, it is not clear which  is the most effective when combined with a vanilla transformer. Second, to our knowledge there is no work addressing the problem of their compatibility with self-supervised methods based on patch masking, and in particular on Bert-like auto-encoders such as BeiT~\cite{bao2021beit}.

In this section we try to answer  these questions. We compare several existing pre-processing designs in terms of accuracy and compute and evaluate them in combination with BeiT, using the  codebase release by the authors of BeiT.
The only change we make is to train the tokenizer on  ImageNet-1k, rather than using the one from DALL-E~\cite{ramesh21arxiv} used in BeiT which is trained on a proprietary dataset comprised of 250 million  images. 
In this manner, pre-training is based on ImageNet-1k only. 
This permits reproducible experimentation and fair comparison, and gives equivalent results~\cite{elnoub2021}. 
Since  existing convolutional designs are not satisfactory in combination with masking, we first introduce our own design. 

\begin{figure}[t]

    \centering
    \includegraphics[width=0.85\linewidth]{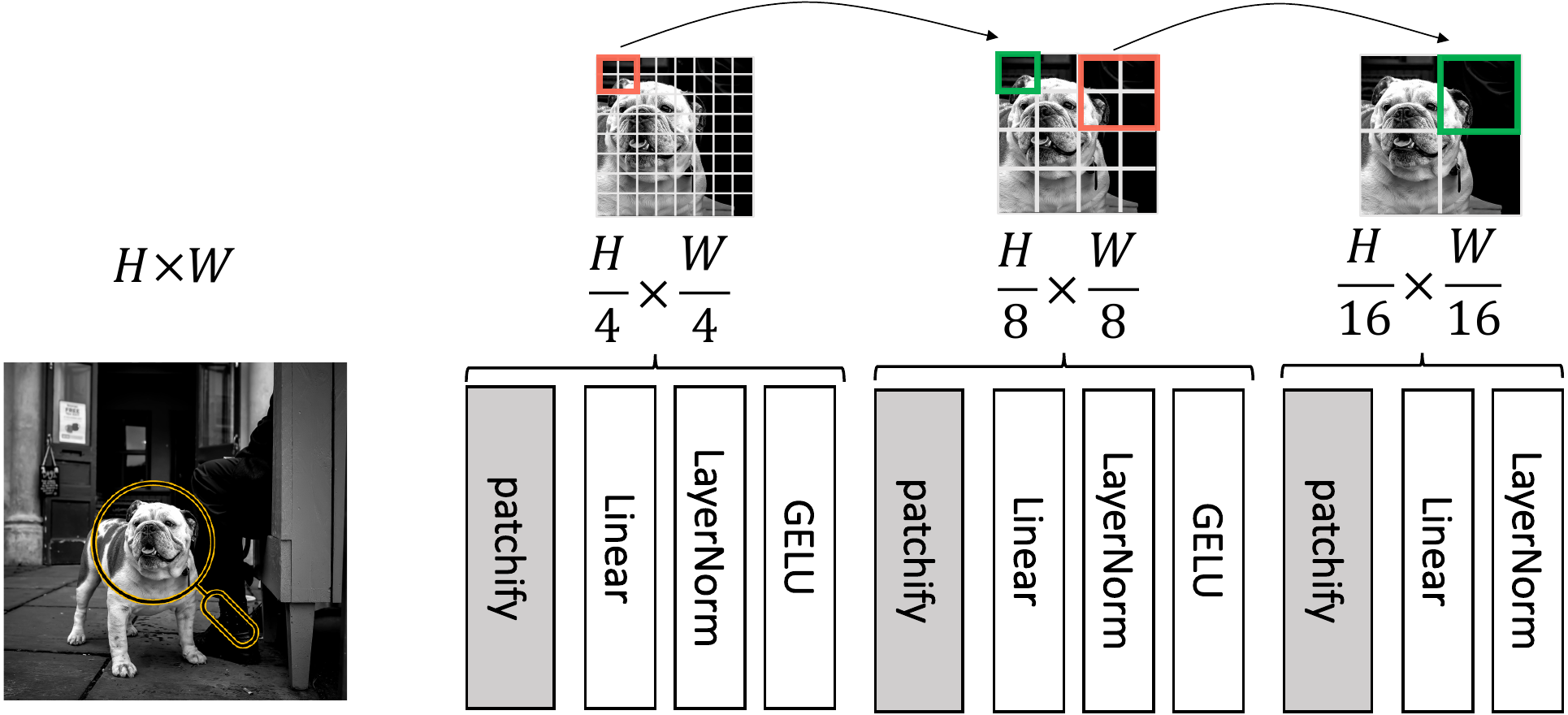} 
    \caption{Design of our hMLP-stem: we start from subpatches and progressively merge them with linear layers interleaved by GELU non-linearities. The design of our stem is such that the patches are processed independently. Hence it commutes with masking.  }
    \label{fig:mlpstem}

\end{figure}

\paragraph{\bf Our hierarchical MLP (hMLP) stem} is depicted in Figure~\ref{fig:mlpstem}. 
All patches are processed independently with linear layers interleaved with non-linearities and renormalization.
Its design is guided by our motivation to remove any interaction between the different 16$\times$16 patches during the pre-processing stage. Even if we mask a patch, it does not create any artifacts resulting from the convolution overlapping with other patches, as it is the case with existing designs. Therefore, with our hMLP solution, we can equivalently mask the patches before or after the patch-processing stage. 
Note that, although patches are processed independently, 
our hMLP-stem is equivalent to a convolutional stem in which the size of the convolutional kernel  and its stride are matched, and in practice we implement it with convolutional layers, see our code in Appendix~\ref{sec:stem_supmat}. 

In short, we start from small 2$\times$2 patches,  and  gradually increase their size until they reach 16$\times$16. Each increase of the patch size is denoted by ``\emph{patchify}'' in Figure~\ref{fig:mlpstem}, in spirit of hierarchical transformer designs like Swin-Transformers~\cite{liu2021swin}. 
The patches are projected with a linear projection and normalized before we apply a GELU non-linearity~\cite{Hendrycks2016GaussianEL}. 
For the normalization, we consider and evaluate two choices: either we use batch-normalization~\cite{Ioffe2015BatchNA} (BN) or layer-normalization (LN)~\cite{ba2016layer}. 
While the BN offers better trade-offs,  LN is of interest when used with small batch sizes: it works well even with a single image per batch, as often used in object detection. 

In contrast with existing stems from the literature, our hMLP design does not significantly increase the compute requirement. For instance,  ViT-B, requires  FLOPS is 17.73 GFLOPS with our design. This  adds less than 1\% of compute compared to using  the usual linear projection stem.

\begin{table}[t]
    \caption{\textbf{Patch pre-processing:} Performance in top1 accuracy with for a ViT-B12. All models are (1) trained 300 epochs in the supervised case; (2) pre-trained during 300 epochs and fine-tuned 100 epochs when used with BeiT. We report the result of a ViT-B13 to provide the performance of a vanilla transformer with more FLOPS. 
    We measure the standard deviation for the two linear  stem baselines and our hMLP stem on 5 runs. The other measurements are made with the fixed seed 0. 
    \label{tab:ablation_stem_design}}
    \centering
\scalebox{0.9}{
\setlength{\tabcolsep}{3pt} 
\begin{tabular}{lcc@{\ }c|lc|l}
\toprule
          &        &     &   &  \multicolumn{2}{c|}{ImNet1k supervised } &  \multicolumn{1}{c}{BeiT+FT}  \\
Stem type & norm. & NL  &  GFLOPS & acc. -val  & acc. -v2 & ImNet-val \\
\midrule
\multirow{4}{*}{Linear: ViT-B12~\cite{dosovitskiy2020image}}  
    & --     & --             & 17.58 &  82.20\stdminus{0.06} & 71.0 & 83.05\stdminus{0.08}  \\ 
    & BN     & --             & 17.58 &  82.31  & 71.0 & 82.98 \\ 
    & --     & GELU           & 17.58 &  81.55 & 70.5 & 83.09 \\ 
    & BN     & GELU           & 17.58 &  82.38 & 70.7 & 82.99 \\
\arrayrulecolor{gray!30} \midrule \arrayrulecolor{black} 
Linear: ViT-B13   & --     & --      &  19.04 & 82.35\stdminus{0.12} & 71.3 &  83.26\stdminus{0.06}\\ 
\midrule
\multirow{2}{*}{Conv: Graham~et al.~\cite{graham2021levit}}  
                                                  & BN  & GELU   & 19.07 & \textbf{82.57} & 71.0 & 83.04  \\ 
                                                  & LN  & GELU   & 19.07 & \textbf{82.50} & 70.9 & 83.06  \\ 
\midrule 
Local transformer ~\cite{Han2021TransformerIT}    &     &        & 19.12 & 82.26 & 70.6 &  82.38 \\ 
\midrule 
\rowcolor{Goldenrod!75}
hMLP (ours)            & BN  & GELU   & 17.73 & \textbf{82.54}\stdminus{0.09} & \textbf{71.5} & \textbf{83.43}\stdminus{0.10}  \\  
& LN  & GELU   & 17.73 & \textbf{82.50}\stdminus{0.07} & 71.0 & 83.24\stdminus{0.09}  \\  
\bottomrule
\end{tabular}}
\end{table}

\mypar{Stem comparison in supervised learning.} 
In Table~\ref{tab:ablation_stem_design} we provide a comparison between different stem designs. We have selected several prototypical designs from the literature for which the code is available online. 
In addition to our hMLP stem, we have considered some variations over the standard linear projection to evaluate the influence of the non-linearities and normalization. 
For the standard linear stem, we also consider a ViT-B13 including an extra pair (MHSA, FFN) to allow more direct comparisons with other stems with more FLOPS. 
In this comparison the most effective existing design is the one of LeViT~\cite{graham2021levit}. 
The improvements with respect to the linear baseline are significant  considering the standard deviation, even when taking into account the extra layer of ViT-B13 to compare with an similar number of FLOPS. 
Our hMLP stem obtains a comparable performance but with lower complexity, and without any interaction between the 16$\times$16 patches. 

\begin{figure}[t]
\begin{minipage}{0.4\linewidth}
\includegraphics[width=\linewidth]{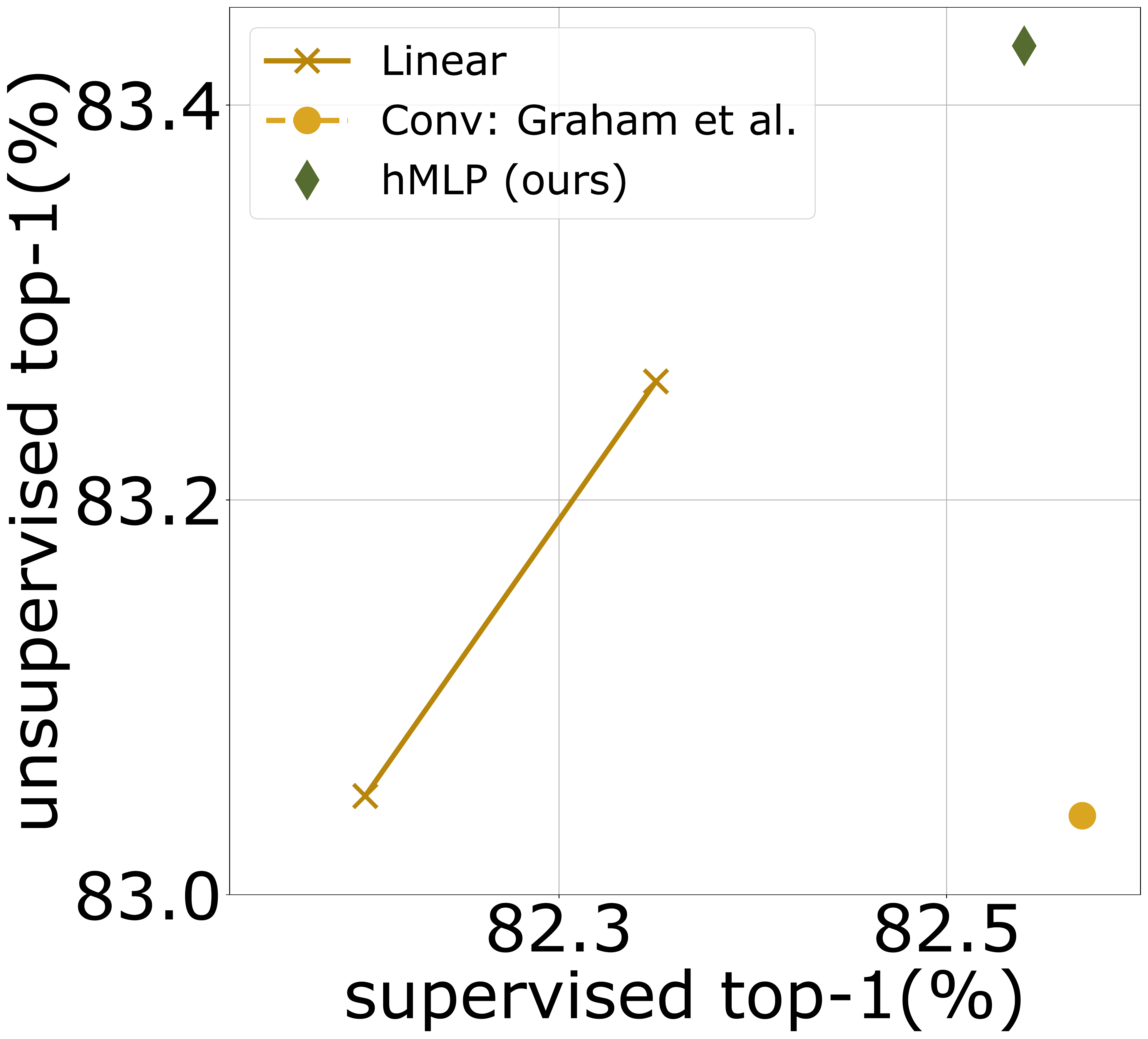}
\end{minipage}
\hfill
\begin{minipage}{0.52\linewidth}
    \caption{Performance of patch pre-processing in the supervised and BeiT+FT settings. Our hMLP stem performs well in both cases, improving the accuracy compared to  linear projection (shown for B12 and B13) without significantly increasing the complexity (+0.8\% FLOPS compared to the ViT-B12 in the bottom-left corner). 
    In contrast, the convolutional stem only improves the performance in the supervised case, while significantly increasing complexity (+7.5\% FLOPS).
    \label{fig:supervised_vs_unsupervised1}}
\end{minipage}
\end{figure}

\mypar{Results with BeiT training.}  
We report the results with BeiT, fine-tuned on ImageNet-val, in the right-most column of Table~\ref{tab:ablation_stem_design}. We use the code of BeiT~\cite{bao2021beit} %
with their  training procedure, which includes LayerScale and a relatively elaborated fine-tuning procedure.  
As one can see, existing stems do not provide any improvement compared to the linear baseline, while adding  compute. In contrast, our design  is effective and provides an improvement of +0.3/+0.4 top1 accuracy compared to the baseline, which is significant considering the measure uncertainty. 
The interest of hMLP in the context of masked self-supervised learning is clear in Figure~\ref{fig:supervised_vs_unsupervised1}, where we plot the performance, averaged over 5 seeds for our method, in the supervised case versus the one with BeiT.

\section{Conclusion}

In this paper, we looked at three different topics related to Vision Transformers. First, we investigated a simple but effective way to parallelize them, showing a viable alternative to increase capacity without significantly increasing the working dimensionality. Whether this simple parallel design principle can be applied to other architectures is an exploration left for future work.
Second, we considered different fine-tuning strategies and showed that fine-tuning the self-attention layer is sufficient in the context of resolution fine-tuning.
This can also be interesting when transferring  to other downstream classification tasks, especially when fine-tuning large models or/and transferring to a dataset with few training images. 
Last, we  introduced a simple patch pre-processing stem, which processes patches independently across multiple linear layers interleaved with non-linearities and patch aggregation.
It is especially useful when combined with mask-based self-supervised learning  such as BeiT. 

\paragraph{Acknowledgement.} We thank Francisco Massa for valuable discussions and insights about optimizing the implementation of block parallelization.

\clearpage
\bibliographystyle{splncs04}
\bibliography{egbib}

\clearpage

\appendix

\begin{center}
{\Large ~\\ \bf Three things everyone should know about ViTs \\ ~\\ -- Supplemental material -- }
\end{center}

\section{Baselines}
\label{sec:baseline_supmat}

\begin{table}[h!]
    \centering
    \setlength{\tabcolsep}{8pt} 
    \scalebox{0.9}{
    \begin{tabular}{lccccl}
         $\downarrow$ Training procedure &  \#epochs &  ViT-Ti & ViT-S & ViT-B & ViT-L \\
         \midrule
         DeiT~\cite{Touvron2020TrainingDI}                     & 300 & 72.2 & 79.8 & 81.8 & \phantom{0.}--  \\
         Steiner et al.~\cite{Steiner2021HowTT}                & 300 & 69.6 & 76.0 & 78.7 & 74.0 \\
         He et al.~\cite{he2021masked}                         & 300 & --   & --   & 82.1 & 81.5$^\dagger$  \\
         He et al.~\cite{he2021masked} with EMA                & 300 & --   & --   & 82.3 & 82.6$^\dagger$  \\
         \rowcolor{Goldenrod}
         Our baseline                                          & 300 & 72.7 & 79.7 & 82.2{\scriptsize$\pm{0.06}$} & 83.0  \\
         \rowcolor{Goldenrod}
         Our baseline with LayerScale~\cite{touvron2021going}  & 400 & 73.5 & 80.7 & 82.7 & 84.0  \\
         \bottomrule
    \end{tabular}}
    \caption{Comparison our baseline with previous training procedures.  We only include results that correspond to the vanilla ViT introduced by Dosovitskiy et al.~\cite{dosovitskiy2020image} for Vit-B, Vit-L and Touvron et al.~\cite{Touvron2020TrainingDI} for  Vit-Ti and ViT-S. All models are  train oned ImageNet-1k at resolution $224 \times 224$ without distillation. $^\dagger$200 epochs.  \label{tab:comp_1k_method}}
\end{table}

\section{Transfer Learning Datasets}
\label{sec:dataset_supmat}

\begin{table}[h!]
\caption{Datasets used in transfer experiments and corresponding references. \label{tab:dataset}}
\centering
    \setlength{\tabcolsep}{8pt} 
\scalebox{0.9}{
\begin{tabular}{l|rrr}
\toprule
Dataset & Train size & Test size & \#classes   \\
\midrule
ImageNet \cite{Russakovsky2015ImageNet12}  & 1,281,167 & 50,000 & 1000  \\ 
iNaturalist 2018~\cite{Horn2017INaturalist}& 437,513   & 24,426 & 8,142 \\ 
iNaturalist 2019~\cite{Horn2017INaturalist}& 265,240   & 3,003  & 1,010  \\ 
Flowers-102~\cite{Nilsback08}& 2,040 & 6,149 & 102  \\ 
Stanford Cars~\cite{Cars2013}& 8,144 & 8,041 & 196  \\  
CIFAR-100~\cite{Krizhevsky2009LearningML}  & 50,000    & 10,000 & 100   \\ 
CIFAR-10~\cite{Krizhevsky2009LearningML}  & 50,000    & 10,000 & 10   \\ 
\bottomrule
\end{tabular}}
\end{table}

\clearpage

\section{Pytorch code of our hMLP Stem}
\label{sec:stem_supmat}

\begin{algorithm}[htb]
\begin{minipage}{1.0\linewidth}
\caption{Pseudocode of hMLP stem in a PyTorch-like style.}
\label{alg:code}
\definecolor{codeblue}{rgb}{0.25,0.5,0.5}
\lstset{
  backgroundcolor=\color{white},
  basicstyle=\fontsize{7.2pt}{7.2pt}\ttfamily\selectfont,
  columns=fullflexible,
  breaklines=true,
  captionpos=b,
  commentstyle=\fontsize{7.2pt}{7.2pt}\color{codeblue},
  keywordstyle=\fontsize{7.2pt}{7.2pt},
}
\begin{lstlisting}[language=python]
import torch 
import torch.nn as nn
class hMLP_stem(nn.Module):
    """ Image to Patch Embedding
    """
    def __init__(self, img_size=(224,224), patch_size=(16,16), in_chans=3, embed_dim=768):
        super().__init__()
        num_patches = (img_size[1] // patch_size[1]) * (img_size[0] // patch_size[0])
        self.img_size = img_size
        self.patch_size = patch_size
        self.num_patches = num_patches
        self.proj = torch.nn.Sequential(
            *[nn.Conv2d(in_chans, embed_dim//4, kernel_size=4, stride=4),
            nn.SyncBatchNorm(embed_dim//4),
            nn.GELU(),
            nn.Conv2d(embed_dim//4, embed_dim//4, kernel_size=2, stride=2),
            nn.SyncBatchNorm(embed_dim//4),
            nn.GELU(),
            nn.Conv2d(embed_dim//4, embed_dim, kernel_size=2, stride=2),
            nn.SyncBatchNorm(embed_dim),
            ])
        
    def forward(self, x):
        B, C, H, W = x.shape
        x = self.proj(x).flatten(2).transpose(1, 2)
        return x
\end{lstlisting}
\end{minipage}
\end{algorithm}

\end{document}